\pdfoutput=1
\documentclass{article}

\usepackage[usenames,dvipsnames,table]{xcolor}
\input{packages/colors.sty}

\PassOptionsToPackage{noabbrev,nameinlink,capitalize}{cleveref}

\usepackage{styles/altpreprint}

\usepackage{natbib}
\usepackage{float}% [H] placement for \texttt{algorithm2e} inside minipages
\usepackage{url}
\usepackage{nicefrac}
\usepackage{enumitem}
\usepackage{needspace}
\usepackage{tikz}
\newsavebox{\sudokuLegendTiedBox}
\newsavebox{\sudokuLegendUntiedBox}
\sbox{\sudokuLegendTiedBox}{\tikz[baseline=0.48ex]{\draw[line width=1pt,color=black!92](0,0)--(6mm,0);}}
\sbox{\sudokuLegendUntiedBox}{\tikz[baseline=0.48ex]{\draw[dash pattern=on 3pt off 2pt,line width=1pt,color=black!85](0,0)--(6mm,0);}}

\newsavebox{\sudokuLegendTiedSideBox}
\newsavebox{\sudokuLegendUntiedSideBox}
\sbox{\sudokuLegendTiedSideBox}{\tikz[baseline=0.42ex]{\draw[line width=0.85pt,color=black!92](0,0)--(4.8mm,0);}}
\sbox{\sudokuLegendUntiedSideBox}{\tikz[baseline=0.42ex]{\draw[dash pattern=on 2.6pt off 2pt,line width=0.85pt,color=black!85](0,0)--(4.8mm,0);}}

\usepackage{packages/algo}
\usepackage{amssymb}
\usepackage{tcolorbox}
\tcbuselibrary{skins}
\usepackage{packages/math}
\usepackage{packages/refs}
\usepackage{packages/symbols}

\usepackage{packages/colab}
\usepackage{appendix}

\crefname{appsec}{appendix}{appendices}
\Crefname{appsec}{Appendix}{Appendices}

\title{Learned Relay Representations for\\ Forward-Thinking Discrete Diffusion Models}

\usepackage[noblocks]{authblk}
\usepackage[symbol]{footmisc}
\renewcommand{\thefootnote}{\fnsymbol{footnote}}
\makeatletter
\renewcommand{\author}[1]{\gdef\@author{#1}}
\makeatother

\author{%
\begin{center}
\mbox{Benjamin Rozonoyer\textsuperscript{1}\thanks{Equal contribution.}\thanks{Corresponding author: \href{mailto:brozonoyer@umass.edu}{brozonoyer@umass.edu}.}~~~~Jacopo Minniti\textsuperscript{2}\footnotemark[1]~~~~Dhruvesh Patel\textsuperscript{1}\footnotemark[1]~~~~Neil Band\textsuperscript{3}}\\Avishek~Joey~Bose\textsuperscript{4,5}~~~~Tim G.\ J.\ Rudner\textsuperscript{2,6}~~~~Andrew McCallum\textsuperscript{1}
\linebreak\linebreak
$^{1}$University of Massachusetts Amherst
\quad
$^{2}$University of Toronto
\\
$^{3}$Stanford University
\quad
$^{4}$Imperial College London
\quad
$^{5}$Mila
\quad
$^{6}$Vijil
\end{center}
}

\date{}

\hypersetup{
  pdftitle={Learned Relay Representations for Forward-Thinking Discrete Diffusion Models},
  pdfsubject={cs.LG, cs.CL},
  pdfauthor={Benjamin Rozonoyer, Jacopo Minniti, Dhruvesh Patel, Neil Band, Joey Bose, Andrew McCallum, Tim G.\ J.\ Rudner},
  pdfkeywords={discrete diffusion, masked language models, backpropagation through time, learned relay representations},
}

\begin{document}

\maketitle
\setcounter{footnote}{0}
\renewcommand{\thefootnote}{\arabic{footnote}}
\vspace{-1.5\baselineskip}
{\centering
  \href{https://www.iesl.cs.umass.edu/diffusion/blog/2026/relay/}{\faIcon{globe}\, Website}
  \qquad
  \href{https://github.com/jacopo-minniti/relay}{\faIcon{github}\, Code}
\par}
\vspace{0.95\baselineskip}

\begin{abstract}
\looseness=-1
When Masked Diffusion Models (MDMs) generate sequences through iterative refinement, the rich internal computation over masked positions is discarded---forcing every subsequent refinement step to recompute the valuable internal information stored as model representations.
To avoid a hard reset between denoising rounds, we propose \OursFull{} (\OursShort{}), a method that allows MDMs to be ``forward-thinking’' when denoising---\emph{explicitly learning how to propagate latent information for the benefit of future denoising steps}. \OursShort{} introduces a differentiable per-token channel that passes information between forward passes and is trained via truncated backpropagation through time (BPTT).
We show that this framework can be scaled to state-of-the-art Diffusion Language Models (DLMs), and is seamlessly compatible with techniques like block diffusion and KV caching.
We first provide a thorough justification of the design choices in \OursShort{} on a challenging Sudoku-based planning task.
We then scale \OursShort~to Fast-dLLM v2, a state-of-the-art DLM, outperforming standard supervised finetuning on coding tasks while reducing the inference latency by up to 32\%.
Our empirical results demonstrate that state-of-the-art DLMs can be explicitly trained to {\it relay} latent information forward across decoding steps, advancing the performance-latency Pareto frontier. We provide code for all our experiments.
\end{abstract}
\vspace{-0.4\baselineskip}

\section{Introduction}
\label{sec:introduction}

\looseness=-1
Masked Diffusion Models (MDMs) generate discrete sequences via iterative denoising~\citep{austin_structured_2021,campbell_continuous_time_2022,sahoo_simple_2024,shi_simplified_2024}: starting from a fully masked canvas, each forward pass unmasks a fraction of the remaining positions. The Transformer computes hidden states at every position---including those still masked---but discards them at the end of each step, beginning the next pass from the partially unmasked sequence alone. We call this the \emph{hard reset} problem: the only information that persists across steps is the discrete tokens just committed, leaving MDMs with no way to accumulate intermediate continuous computation.

\looseness=-1
This matters because recurrent computation---unrolling a fixed-parameter model across many steps---is exactly the structural property that recent work has tied to improved performance on difficult reasoning tasks, as it effectively expands the function class the model can approximate~\citep{pmlr-v235-gatmiry24b, saunshi2025reasoning, li2024chain}. MDMs already perform many forward passes per generation; the hard reset is what prevents any of that compute from being reused.

\vspace*{3pt}
% \vfill
\begin{tcolorbox}[
  enhanced,
  colback=blue!4,
  colframe=blue!55!black,
  leftrule=3pt,
  toprule=0.4pt,
  bottomrule=0.4pt,
  rightrule=0.4pt,
  arc=2pt,
  boxsep=1.5pt,
  left=8pt, right=8pt, top=4pt, bottom=4pt,
  width=\linewidth,
  before skip=2pt,
  after skip=5pt,
  % fontupper=\small,
]
\centering
{\bfseries This raises a natural question}:
How can the sequential unmasking structure of MDMs support recurrent computation that carries richer information across steps?
\end{tcolorbox}

\pagebreak

\looseness=-1
Our answer is \OursFull{} (\OursShort{}), a method that makes discrete diffusion models \emph{forward-thinking}: at each denoising step, alongside any newly unmasked tokens, the model carries its last-layer hidden states forward as a learned \emph{relay}, giving the next forward pass direct access to the prior step's continuous computation. Simply piping these states forward, however, does not by itself ensure they encode anything useful for what follows. \OursShort{} therefore trains the relay end-to-end with truncated backpropagation through time \citep[BPTT;][]{werbos_bptt_1990}, shaping it to be maximally informative for the next several denoising steps and enabling a form of latent chain-of-thought across the unmasking trajectory.

\looseness=-1
\xhdr{Contributions}
We introduce \OursShort{}, which equips MDMs with learned relay representations---continuous latent states passed forward across decoding steps and trained end-to-end via truncated BPTT. \OursShort{} is architecture-agnostic and leaves the inference-time decoding procedure of MDMs (unmasking schedule, sampling) unchanged; the only addition at inference is forwarding the relay alongside the committed tokens. It is also compatible with prevalent DLM acceleration techniques, including block diffusion~\citep{arriola_block_2024} and KV caching~\citep{wu2025fastdllm,wu2025fastdllmv2efficientblockdiffusion}.

To summarize, our key contributions are as follows:\vspace*{-4pt}
\begin{enumerate}[topsep=2pt, align=left, leftmargin=15pt, labelindent=1pt,
listparindent=\parindent, labelwidth=0pt, itemindent=!, itemsep=2pt, parsep=0pt]
    \item \looseness=-1 We propose \OursShort{}, a general method for incorporating recurrent computation in MDMs by training the model---via truncated BPTT---to pass a learned latent relay forward across decoding steps. \OursShort{} can train an MDM from scratch or adapt a pre-trained MDM through lightweight adaptation.
    
    \item \looseness=-1 We validate \OursShort{} at LLM scale through full-parameter adaptation of Fast-dLLM~v2 1.5B~\citep{wu2025fastdllmv2efficientblockdiffusion}, outperforming standard supervised finetuning on coding tasks while reducing inference latency by up to 32\%.
    
    \item \looseness=-1 We perform extensive ablations that map out the design space of \OursShort{} and validate our choices.
\end{enumerate}

\section{Background: Masked Diffusion Models}\label{sec:background}
We tackle the hard reset problem in masked diffusion models by training them to pass along a learned relay state. Before presenting our approach, \OursShort{}, we review the training and inference procedure for Masked Diffusion Models (MDMs)~\citep{shi_simplified_2024, sahoo_simple_2024} that \OursShort{} builds upon.

\xhdr{Notation} 
We denote the vocabulary as $\gV$, including the $\mask{}$ token. The space of sequences of length $L$ over the vocabulary is $\gV^L$.
Superscripts denote the position in the sequence, e.g., $x^i$ is the $i$-th token in the sequence $\vx \in \gV^L$.
$\mathcal M(\vx)\subseteq [L]$ denotes the set of masked positions in the sequence $\vx$.

\xhdr{Training} 
The noising process proceeds by sampling a time $t\in [0, 1]$ and masking each position in a clean sequence $\vx_0\in (\gV \setminus \{\mask\})^L$ independently with probability $\alpha_t$, to obtain the noised (partially masked) sequence $\vx_t$.
The coordinate-wise posterior distribution $\sP(X_0^i=x_0^i \,|\, \X_t=\vx_t)$ is denoted as $p(x_0^i \mid \vx_t)$. 
As noted in \citet{zheng_time_agnostic_masked_2024}, this posterior depends on $\vx_t$ only through its masked pattern and revealed tokens, not on the time $t$ itself.
The coordinate-wise posterior is parameterized by a neural network denoted as $p_\theta^i(\cdot \mid \vx_t) \in \Delta$ for $i \in \mathcal M(\vx_t)$ and is trained by minimizing the weighted sum of cross-entropy losses for each masked position\footnote{We have assumed a linear noise schedule.}
%consisting of a backbone $\fTheta$ and a projection head $\unemb$, which are :
\begin{equation}\label{eq:mdm_training_objective}
    \gL(\theta) = \E_{\vx_0, t, \vx_t}\left[\frac{1}{t} \sum\nolimits_{i: \vx_t^i=\mask} - \log p^i_\theta(x_0^i \mid \vx_t)\right].
    %~~~\text{where }~ p_\theta(x_0^i \mid x_t) = \frac{e^{\fTheta(\unemb(\x_0))^i}}{\sum_{v \in \gV} e^{\fTheta(\unemb(\vx_t))^i}}.
\end{equation}
%where we have assumed a linear noise schedule.
The coordinate-wise parametric posterior is implemented using embedding $\emb: \gV \to \mathbb R^d$, unembedding $\unemb: \mathbb R^d \to \mathbb R^{|\gV|}$, and a transformer backbone $\fTheta: \gV^L \to \mathbb R^{L\times d}$
that produce the posterior distribution:
\begin{align*}
  p_\theta^i(w \mid \vx_t)
  &= \frac{e^{\ell^i(w)}}{\sum_{w' \in \gV} e^{\ell^i(w')}},\quad
  \text{where}\quad
  \ell^i(w) = \unemb(\fTheta(\emb(\vx_t)))^i_w.
\end{align*}

\xhdr{Inference}
Generation proceeds along a decreasing time grid $1=t_0 > t_1 > \cdots > t_K=0$, iteratively unmasking positions from the all-masked sequence $\vx_{t_0}=(\mask,\ldots,\mask)$ to a fully unmasked sequence $\vx_{t_K}\in(\gV \setminus \{\mask\})^L$.
At each step $k$, given the current partially masked sequence $\vx_{t_k}$, the model computes logits $\bm{\ell}_k$ for the per-position posterior distribution for each masked position $i \in \mathcal M(\vx_{t_k})$ and token $w \in \gV$.
An unmasking policy $u(\cdot \mid \bm{\ell}_k, \vx_{t_k})$, which may be stochastic, then selects a set of positions $\mathcal U_k \subseteq \mathcal M(\vx_{t_k})$ to reveal, producing the next partially masked sequence $\vx_{t_{k+1}}$.
Common choices for $u(\cdot \mid \bm{\ell}_k, \vx_{t_k})$ include unmasking a fixed fraction of the remaining masks at each step~\citep{nie_llada_2025} and confidence-based rules~\citep{ben-hamuAcceleratedSamplingMasked2025, kim_train_worst_plan_best_2025,patel2025improved}.

\xhdr{The Hard Reset Problem}
After each inference step, MDMs discard the entire computational state used to choose the newly revealed tokens. 
The next step starts again from $\vx_{t_{k+1}}$ alone. 
%Thus, information can be transmitted from one iteration to the next only through newly committed tokens. 
Thus, standard MDM inference treats each partially masked sequence as a fresh prediction problem---a \emph{hard reset}---rather than as a continuation of an ongoing computation. 
Because models can only perform a constant number of FLOPs in each forward pass, hard reset prevents the model from amortizing reasoning across steps effectively.
In the next section, we propose our solution to this problem: we learn a continuous latent state that is passed across the steps of MDM inference.

\section{\OursFull{}}\label{sec:method}
To address the \emph{hard reset} problem, we introduce a continuous differentiable state that is carried across MDM inference steps and can circumvent the hard reset.

\subsection{Augmented State Trajectories}
The training of MDMs proceeds by sampling a data point $\vx_0\sim p_{\mathrm{data}}$, a time $t \sim \mathcal{U}(0, 1)$, and a partially masked sequence $\vx_t$ under the noise schedule given the time $t$ and the data point $\vx_0$.
During inference, we have a discretized time grid $1=t_0 > \cdots > t_n = 0$, and the corresponding inference trajectory $\vx_{t_0}, \ldots, \vx_{t_n}$ obtained by using some unmasking policy $u$, where $\vx_{t_0}=\{\mask\}^L$.
We wish to pass a continuous state forward across decoding steps, that can carry intermediate computations from the previous step which have not yet been realized as a decoded token. 
We can break down this behavior into two primitives: a model must produce a relay state $\vh_k$ at inference step $k$, and learn to consume that relay state at step $k{+}1$.
\Cref{fig:relay_schematic} shows a schematic of the augmented state trajectory produced by the model, where $\vs_k = (\vx_{t_k}, \vh_k)$ is the augmented state at step $k$.

\subsection{Training}

\xhdr{Architecture}
We parameterize the augmented dynamics with a backbone $\fTheta$, relay module $\Rphi$, token embedding $\emb$, and unembedding head $\unemb$ (see \Cref{fig:relay_schematic}).
At step $k$, the model maps the current pair $(\vx_{t_k}, \vh_k)$ to the next relay state and per-position logits via
\begin{align}
  \vh_{k+1} &= \fTheta\!\left(\emb(\vx_{t_k}) + \Rphi(\vh_k)\right),
  &
  \bm{\ell}_k &= \unemb(\vh_{k+1}),
  \label{eq:relay_recurrence}
\end{align}
initialized with $\vh_0 = \vzero$. The per-position posterior $p_\theta^i(\cdot \mid \vx_{t_k}, \vh_k)$ is read off from $\bm{\ell}_k$ by a softmax, exactly as in standard MDMs.

Since we only care about the terminal state $\vx_{t_n}$, we continue to provide supervision using the same cross-entropy loss as in standard MDMs, and train the model to produce useful relay states $\vh_k$ that help improve predictions $K$ steps ahead using truncated BPTT.
Specifically, instead of sampling $\vx_t$ as in standard MDMs, we start from an all masked sequence $\vx_{t_0}=\{\mask\}^L$ and roll out under \Cref{eq:relay_recurrence} together with an unmasking policy $u$ (see below), producing the augmented trajectory $(\vx_{t_0}, \vh_0), \ldots, (\vx_{t_n}, \vh_n)$. The total training loss is the expected sum of per-step cross-entropies over the trajectory:
\begin{align}
  \mathcal{L}(\theta)
  &= \E_{\vx_0,\, \xi_{0:n-1}}\!\left[\,\sum_{k=0}^{n-1}\,\sum_{i \in \mathcal{M}(\vx_{t_k})} -\log p^i_\theta\!\left(x_0^i \mid \vx_{t_k}, \vh_k\right)\right],
  \label{eq:relay_training_objective}
\end{align}
where $\xi_{0:n-1}$ denotes the exogenous randomness used by the unmasking policy along the rollout. Unlike an externally observed conditioning variable, $\vh_k$ is an internal artifact of the rollout, part of the computational trajectory rather than of the generated object. At inference time each step carries forward the realized pair $(\vx_{t_k}, \vh_k)$, but only $\vx_{t_k}$ is eventually decoded into text, while $\vh_k$ serves as a differentiable memory channel for future predictions. The full procedure is summarized in \Cref{alg:training}; we derive the gradient estimator below.
\begin{figure}[t]
  \centering
  \captionsetup{font=footnotesize,skip=4pt}
  \begin{minipage}[t]{0.46\textwidth}
    \vspace{0pt}%
    \footnotesize
    \begin{algorithm2e}[H]
      \LinesNumbered
      \caption{\OursShort{} Training}
      \label{alg:training}
      \KwIn{model $\fTheta$, relay module $\textcolor{relayink}{\Rphi}$, unroll horizon $K$, unmasking policy $u$, training steps $N$, learning rate $\eta$}
      \KwSty{for}\ $t \in \{1, \ldots, N\}$\ \KwSty{do}\;
      \Indp
        \AlgoBlockGap
        \AlgoShade{algreset}%
        \KwSty{if}\ $t=1$ \textnormal{or} $\mathcal{M}(\vz) = \emptyset$\ \KwSty{then}\;
        \Indp
          $\vx_0 \sim p_{\mathrm{data}},~ \vz \gets \{\mask\}^L,~ \textcolor{relayink}{\vh \gets \vzero}$\;
        \Indm
        \KwSty{end}\;
        \AlgoBlockGap
        \AlgoShade{algroll}%
        $L \gets 0$\;
        \KwSty{for}\ $k \in \{0, \ldots, K-1\}$\ \KwSty{do}\;
        \Indp
          $\textcolor{relayink}{\vh} \gets \fTheta\!\left(\emb(\vz) + \textcolor{relayink}{\Rphi(\vh)}\right)$\;
          $\bm{\ell} \gets \unemb(\textcolor{relayink}{\vh})$\;
          $L \gets L + \mathcal{L}(\bm{\ell}, \vx_0)$\CommentFit{masked only}
          $\mathcal{U} \sim u(\cdot \mid \bm{\ell}, \vz)$\;
          $\textcolor{relayink}{z^i \gets x_0^i} ~~ \forall i \in \mathcal{U}$\;
        \Indm
        \KwSty{end}\;
        \AlgoBlockGap
        \AlgoShade{algstep}%
        $\theta \gets \theta - \eta \, \nabla_{\theta} L$\;
        \AlgoNoShade
        \AlgoBlockGap
      \Indm
      \KwSty{end}\;
      \Return{$\theta$}
    \end{algorithm2e}
    \vspace{-0.15em}
    {\centering\scriptsize
      \AlgoLegend{algreset}{reset}\hspace{0.85em}%
      \AlgoLegend{algroll}{rollout}\hspace{0.85em}%
      \AlgoLegend{algstep}{update}\par}
  \end{minipage}\hfill
  \parbox[t]{0.50\textwidth}{%
    \vspace{0pt}%
    \centering
    \resizebox{0.88\linewidth}{!}{%
      \input{figures/relay_schematic.tex}%
    }\par\smallskip
    \begingroup
      \captionsetup{%
        font=footnotesize,%
        skip=4pt,%
        margin={0pt,0pt},%
        width=0.94\linewidth,%
        hypcap=false,%
      }%
      \captionof{figure}{Schematic of \OursShort{} over two consecutive inference steps.  At each step $k$, the backbone $\fTheta$ consumes the sum of embedded tokens $\emb(\vx_{t_k})$ and the projected relay state $\Rphi(\vh_k)$, producing a hidden state $\vh_{k+1}$ that is both unembedded into logits for the cross-entropy loss and forwarded through the relay module $\Rphi$ (\textcolor{relay}{orange} path) into the next step. Tokens are progressively unmasked between steps (newly committed tokens {\setlength{\fboxsep}{0.6pt}\colorbox{relay!40}{highlighted}}; e.g.\ $\mask\!\to\!$\,f at step $k$, $\mask\!\to\!$\,b,\,c at step $k{+}1$), while $\vh$ provides a continuous, differentiable channel for information that has not yet been committed to a discrete token.}%
      \label{fig:relay_schematic}%
    \endgroup
  }
\end{figure}

\looseness=-1
\xhdr{Constructing rollouts}
In order to perform truncated BPTT, we need to construct rollouts of the augmented state trajectory under an unmasking policy $u$. Given the current augmented state $(\vx_{t_k}, \vh_k)$, one step of rollout proceeds as follows:\vspace*{-2pt}
\begin{itemize}[nosep, leftmargin=24pt, labelindent=0pt]
  \item \textbf{Position selection:} Sample which positions to unmask, $\gU \sim u(\cdot \mid \bm{\ell}_k,\vx_{t_k})$. The policy may use the model's own logits $\bm{\ell}_k$.
  \item \textbf{Token forcing:} For each $i \in \mathcal{U}$, commit the token from ground truth: $x_{t_{k+1}}^i = x_0^i$.
\end{itemize}
We teacher-force the token \emph{values} (rather than sampling from the model's posterior $p_\theta^i(\cdot \mid \vx_{t_k}, \vh_k)$) because sampled values would inject errors that the rollout has no mechanism to correct.
The \emph{position} sampler, by contrast, may use the model's own posterior without affecting the ideal minimizer: in absence of the continuous channel this leaves the standard MDM training objective (\Cref{eq:mdm_training_objective}) unchanged~\citep{kim_puma_2026}, and for the augmented-state trajectory the same argument applies but a formal proof requires additional assumptions and is more involved.
%and left to future work.
%We provide formal statements in \Cref{sec:proofs}.

\looseness=-1
\xhdr{Gradient estimation}
We now derive the gradient estimator for one $K$-step window of the recurrence \Cref{eq:relay_recurrence}. Let $\xi_k$ denote the exogenous randomness used in the sampled unmasking step at $k$:
\begin{align}
  \gU_k
  &\sim
  u(\cdot \mid \bm{\ell}_k, \vx_{t_k}),
  \quad\text{and}\quad
  x_{t_{k+1}}^i \gets x_0^i
  \quad \forall i \in \gU_k.
\end{align}
Conditioning on the realized $\xi_{0:K-1}$, the per-window loss is
\begin{align}
  \mathcal{L}_K(\theta; \vx_0, \xi_{0:K-1})
  &=
  \sum_{k=0}^{K-1}
  L_k(\bm{\ell}_k, \vx_0),
  \label{eq:relay_rollout_objective}
\end{align}
where $L_k$ is the per-step cross-entropy at step $k$ (summed over the masked positions of $\vx_{t_k}$), and $\bm{\ell}_k$, $\vh_{k+1}$ are computed from $(\vx_{t_k}, \vh_k)$ via \Cref{eq:relay_recurrence}.
The discrete update $\vx_{t_k} \to \vx_{t_{k+1}}$ is treated as fixed after the rollout is sampled.
Equivalently, this estimator sets $\partial \vx_{t_{k+1}} / \partial \bm{\ell}_k = 0$ and does not differentiate through the sampled unmasking decisions.
The BPTT adjoints over the differentiable relay state are then defined by
\begin{align}
  \lambda_K &= 0, \nonumber\\
  \lambda_k
  &=
  \left(
    \partial_{\vh_k} \bm{\ell}_k
  \right)^\top
  \nabla_{\bm{\ell}_k}L_k(\bm{\ell}_k, \vx_0)
  +
  \left(
    \partial_{\vh_k} \vh_{k+1}
  \right)^\top
  \lambda_{k+1},
  \qquad k=K-1,\ldots,0.
  \label{eq:relay_adjoint_recursion}
\end{align}
Throughout, $\partial_{\vh_k}\bm{\ell}_k$ and $\partial_\theta\bm{\ell}_k$ denote the \emph{total} derivatives along the single-step chain $\vh_k \to \vh_{k+1} \to \bm{\ell}_k$, i.e., $\partial_{\vh_k}\bm{\ell}_k = (\partial_{\vh_{k+1}}\unemb)(\partial_{\vh_k}\vh_{k+1})$, and analogously for $\theta$; the companion factor $(\partial_\theta \vh_{k+1})^\top \lambda_{k+1}$ below uses the \emph{direct} partial of step $k$'s transition only ($\vh_k$ held fixed). The boundary $\lambda_K=0$ therefore reads as ``no downstream losses past step $K{-}1$.''
The corresponding sampled gradient estimator is
\begin{align}
  \nabla_\theta \mathcal{L}_K
  &=
  \sum_{k=0}^{K-1}
  \left[
  \begin{aligned}
    &\underbrace{
    \left(
      \partial_\theta \bm{\ell}_k
    \right)^\top
    \nabla_{\bm{\ell}_k}L_k(\bm{\ell}_k, \vx_0)
    }_{\text{direct gradient from immediate cross-entropy}}
    +
    \underbrace{
    \left(
      \partial_\theta \vh_{k+1}
    \right)^\top
    \lambda_{k+1}
    }_{\text{BPTT through relay state}}
  \end{aligned}
  \right].
  \label{eq:relay_bptt_gradient}
\end{align}
For a two-step truncation beginning at step $k$, we have $\lambda_{k+2}=0$, so the only downstream adjoint is
\begin{align}
  \lambda_{k+1}
  =
  \left(
    \partial_{\vh_{k+1}} \bm{\ell}_{k+1}
  \right)^\top
  \nabla_{\bm{\ell}_{k+1}} L_{k+1}(\bm{\ell}_{k+1}, \vx_0).
\end{align}
The two-step gradient is therefore
\begin{align}
  \nabla_\theta \left(L_k + L_{k+1}\right)
  &=
  \underbrace{
  \sum_{j=k}^{k+1}
  \left(
    \partial_\theta \bm{\ell}_j
  \right)^\top
  \nabla_{\bm{\ell}_j} L_j(\bm{\ell}_j, \vx_0)
  }_{\text{direct gradient from immediate cross-entropy}}
  \nonumber
  +
  \underbrace{
  \left(
    \partial_\theta \vh_{k+1}
  \right)^\top
  \lambda_{k+1}
  }_{\text{BPTT through relay state}}
  .
  \label{eq:relay_two_step_gradient}
\end{align}

Thus, each step receives the local cross-entropy gradient through its logits $\bm{\ell}_k$, and the additional recurrent gradient is back-propagated through the differentiable relay path $\vh_k\rightarrow \vh_{k+1}$.
\section{Experiments}\label{sec:experiments}

Through our experiments we seek to address the following research questions:
\begin{description}[font=\bfseries, leftmargin=1.5cm, labelindent=16pt]
    \item[RQ1] Does training to be forward-thinking with BPTT improve performance and latency?
    \item[RQ2] Does weight-tying $\emb$ and $\unemb$ have an impact on \OursShort{}, since $f_\theta$ at the first layer must learn to consume the $\unemb$-aligned relay $\vh$ from the last layer?
    \item[RQ3] Can we efficiently adapt state-of-the-art DLMs to use relay representations and improve their performance-latency frontiers?
\end{description}

We first motivate the design choices for \OursShort{} with a thorough ablation on Sudoku. Subsequently, we post-train Fast-dLLM v2 \citep{wu2025fastdllmv2efficientblockdiffusion}, a state-of-the-art DLM, demonstrating the effectiveness of \OursShort{} on model adaptation for DLMs.

\subsection{Sudoku}\label{sec:sudoku}

\looseness=-1
\xhdr{Dataset} The objective of a Sudoku puzzle is to fill in a 9x9 board (of nine 3x3 sub-squares) with digits 1-9 such that each row, column, and 3x3 square contains all the nine unique digits. A puzzle has a minimum of 17 clues, which is a mathematical prerequisite for it to have a unique solution \citep{mcguire2014there}.

\looseness=-1
\xhdr{Setup} We choose the Sudoku-Extreme dataset \citep{wang2025hierarchical} as a challenging benchmark that allows us to focus on modeling choices without the risk of overfitting. We release a derived version\footnote{\url{https://huggingface.co/datasets/brozonoyer/sapientinc-sudoku-extreme-timvink-sudoku-solver}} that augments each puzzle with a step-by-step solver trajectory, step count, and the set of deduction strategies invoked, obtained by running the Sudoku solver of \citet{vink2024sudokusolver}\footnote{\url{https://github.com/timvink/sudoku-solver}} over every example; the ``strategies'' used by the solver underpins the \emph{deduction-only} evaluation slice in \Cref{tab:sudoku-tau015}. For the experiments in \Cref{fig:sudoku-pareto} and \Cref{tab:sudoku-tau015} we evaluate on the first 2k examples of the test split. All methods use the same small Transformer backbone ($\sim 7$M parameters; full architecture in \Cref{sec:appendix_details}) with rotary position embeddings and are trained to convergence, in line with our experimental protocol of comparing methods by their test-time performance versus latency frontiers.
\footnote{We use the xLM package (\url{https://github.com/dhruvdcoder/xlm-core}) for all small-scale experiments on Sudoku.}
Our predictor uses top-probabilities as confidence values $c_i$, sorts by increasing $1-c_i$, and unmasks all positions whose cumulative confidence falls below a threshold $\tau$, falling back on the argmax if no such position exists. For \OursShort{}'s on-policy training rollout we use a stochastic threshold $\tau \sim \gN(\mu=0.15, \sigma=0.1)$ for robustness (the threshold is a hyperparameter of the sampling decision $\gU \sim u(\cdot \mid \bm{\ell}, \vz)$ in \Cref{alg:training} line 10).

\looseness=-1
\xhdr{Baselines and ablations} We compare four training objectives that progressively turn on the components of \Cref{alg:training}, each instantiated with both \emph{tied} and \emph{untied} embeddings (whether $\emb$ and $\unemb$ share weights). \textbf{MLM} \citep{sahoo_simple_2024,shi_simplified_2024} is standard uniform masked diffusion: a single forward pass per training step ($K{=}1$), no relay ($R_\theta\!\equiv\!0$, so $\vh \gets \fTheta(\emb(\vz))$), and no inner rollout. Instead, the masked input $\vz$ is drawn fresh each step by sampling $t \sim \mathcal{U}(0,1)$ and masking each token of $\vx_0$ independently with probability $t$. The remaining three objectives all share \OursShort{}'s on-policy \emph{position} sampler $u(\cdot\mid\bm\ell,\vz)$ (\Cref{alg:training} line 10) and teacher-force the committed positions to the values in $\vx_0$ between passes (line 11), differing only in whether and how the relay channel is used (a related rollout training procedure is studied by \citealp{kim_puma_2026}). \textbf{Rollout} unrolls $K{=}2$ inner steps but keeps $R_\theta\!\equiv\!0$ so each step recomputes $\vh$ from $\emb(\vz)$ alone; this isolates the contribution of \emph{which} positions get committed between forward passes. \textbf{\OursShort{} (sg)} additionally enables the relay path $R_\theta(\vh)$ inside the inner loop but stop-gradients $\vh$ before feeding it back, so the backbone receives no temporal credit across the $K$ steps. \textbf{\OursShort{}} is the full method: $K{=}2$ BPTT through the relay (\Cref{alg:training}). At inference we sweep deterministic thresholds $\tau \in \{0.05, 0.10, 0.15, 0.20, 0.25\}$ and trace each method's accuracy-NFE frontier; lower $\tau$ commits fewer cells per forward pass and so spends more NFEs.

\begingroup
\setlength{\intextsep}{6pt plus 2pt minus 2pt}%
\begin{figure*}[t]
\centering
\includegraphics[width=\textwidth]{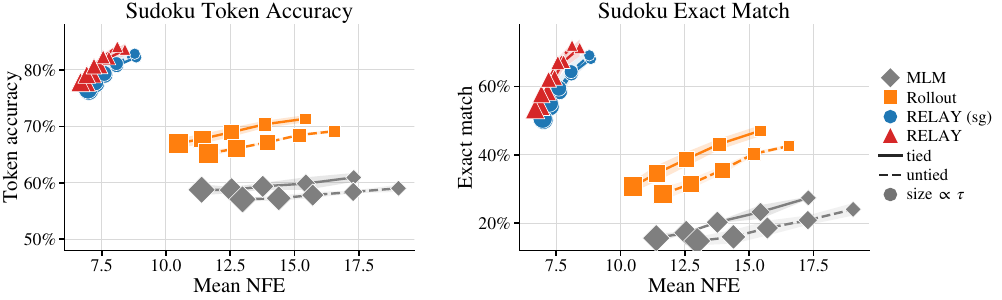}
\caption{Accuracy-NFE frontier on Sudoku-Extreme test. Each curve traces a single training method as we sweep the inference confidence threshold $\tau \in \{0.05, 0.10, 0.15, 0.20, 0.25\}$. A lower $\tau$ commits fewer cells per forward pass and so spends more NFEs (rightward), and vice-versa. Shaded ribbons denote $\pm 1$ sample standard deviation across three training seeds within each tying condition.}
\label{fig:sudoku-pareto}
\end{figure*}
\endgroup

\xhdr{Carry-mechanism ablation}
The additive carry in \Cref{eq:relay_recurrence} is itself a design choice, which we validated during development by comparing three mechanisms for passing latent state between denoising steps: (i) \OursShort{}'s additive carry, (ii) a bottleneck MLP bridge, and (iii) a bottleneck cross-attention bridge in which the current step's token embeddings cross-attend to the previous step's hidden states. For a single-seed ablation run with BPTT through the recurrence, exact match at $\tau{=}0.15$ reached $65.1\%$ (additive), $62.3\%$ (MLP bridge), and $46.9\%$ (cross-attention bridge), versus $60.4\%$ for the additive carry with stop-gradient, so we adopt the simple additive carry throughout. We also ablated the BPTT unroll horizon and found $K{=}2$ to be optimal: longer horizons ($K{\in}\{3,4\}$) do not improve accuracy and make optimization less reliable (\Cref{sec:bptt_horizon}).

\looseness=-1
\xhdr{Results and analysis} \Cref{fig:sudoku-pareto} plots test metrics at the latest checkpoint for each seed. Replacing uniform masking (\textbf{MLM}) with the on-policy confidence-thresholded sampler under teacher forcing of the unmasked values (\textbf{Rollout}) yields the first improvement. 
%\footnote{Preliminary experiments without teacher-forcing intoduced exposure bias and degraded performance significantly, consistently with the observations of \citet{kim_puma_2026}.}
Turning the relay channel on (\textbf{\OursShort{} (sg)}) contributes the next big jump, highlighting the importance of a soft state carried between forward passes. Finally, replacing the stop-gradient with $K{=}2$ BPTT through the relay (\textbf{\OursShort{}}, \Cref{alg:training}) yields a further separation and the best accuracy-NFE frontier across thresholds.%\TODO{for \OursShort+StopGrad+tied+seed1 and for MLM+uniform+tied+seed3 we take eval steps 285k and 20k, respectively}

\looseness=-1
We are able to trace this last separation of \OursShort{} over \OursShort{} (sg) to the fact that, at the same threshold $\tau$, \OursShort{} \textbf{commits more cells per forward pass while keeping the partial board legal} --- where a board is \emph{legal} when no row, column, or 3$\times$3 box yet contains a repeated digit. Legality is a necessary condition for correctness, and is well-defined at every intermediate denoising step, not only at the end. Since the studied architectures cannot perform recursive search, we restrict this qualitative analysis to a \emph{deduction-only} cohort of 2{,}000 test puzzles for which the solver uses only human-like deduction strategies (Advanced or Master heuristics; cohort construction detailed in \Cref{sec:dataset_details}).

\looseness=-1
At the matched threshold $\tau=0.15$, \OursShort{} produces a fully legal final board $74.8\%$ of the time versus $70.7\%$ for \OursShort{} (sg) ($+4.1$ pp), and incurs $15\%$ fewer row/column/box violations across the rollout ($0.90$ vs.\ $1.06$ on average per puzzle); these legality gains are uniform across the Advanced ($+4.0$ pp) and Master ($+4.1$ pp) strata. In other words, \textbf{BPTT teaches the relay to keep the partial board self-consistent under more aggressive unmasking}: at the same confidence threshold $\tau$, \OursShort{} commits more cells per forward pass while still honoring the row/column/box constraints, so the rollout reaches the same accuracy in fewer total forward passes --- producing the strict outward shift of the $(\tau \to \text{accuracy-NFE})$ frontier in \Cref{fig:sudoku-pareto}. \Cref{tab:sudoku-tau015} reports the corresponding exact match and mean NFE at $\tau=0.15$ on both the unfiltered test split and the deduction-only cohort: \OursShort{} attains the highest exact match and the lowest mean NFE in every (slice, tying) cell, with up to a $+5.7$ pp gain over \OursShort{} (sg) in the tied setting at uniformly lower NFE. Tying $\emb$ and $\unemb$ yields a marginal improvement for \OursShort{} ($\leq 3.5$ pp exact match across all rows), consistent with the small Sudoku vocabulary leaving the residual stream ample capacity to carry both predictive and relay-bearing information.

\begin{table*}[t]
  \centering
  {\footnotesize
  \setlength{\tabcolsep}{14pt}
  \begin{tabular}{@{}lc r@{.}l r@{.}l r@{.}l r@{.}l@{}}
    \toprule
    \textbf{Objective} & \textbf{Tying} &
    \multicolumn{2}{c}{\textbf{\shortstack{Acc (\%)\\unfiltered}}} &
    \multicolumn{2}{c}{\textbf{\shortstack{Mean NFE\\unfiltered}}} &
    \multicolumn{2}{c}{\textbf{\shortstack{Acc (\%)\\deduction-only}}} &
    \multicolumn{2}{c}{\textbf{\shortstack{Mean NFE\\deduction-only}}} \\
    \midrule
    \multirow{2}{*}{MLM} & $\checkmark$ & 20 & 27\%{\scriptsize $\pm$\,0.25\%} & 13 & 76{\scriptsize $\pm$\,0.56} & 32 & 23\%{\scriptsize $\pm$\,1.87\%} & 9 & 27{\scriptsize $\pm$\,0.17} \\
    & $\times$ & 18 & 58\%{\scriptsize $\pm$\,2.17\%} & 15 & 70{\scriptsize $\pm$\,0.32} & 35 & 67\%{\scriptsize $\pm$\,2.67\%} & 10 & 23{\scriptsize $\pm$\,0.31} \\
    \midrule
    \multirow{2}{*}{Rollout} & $\checkmark$ & 38 & 70\%{\scriptsize $\pm$\,2.09\%} & 12 & 54{\scriptsize $\pm$\,0.31} & 52 & 93\%{\scriptsize $\pm$\,3.08\%} & 10 & 45{\scriptsize $\pm$\,0.33} \\
    & $\times$ & 35 & 55\%{\scriptsize $\pm$\,1.15\%} & 13 & 94{\scriptsize $\pm$\,0.46} & 54 & 35\%{\scriptsize $\pm$\,4.32\%} & 11 & 13{\scriptsize $\pm$\,0.23} \\
    \midrule
    \multirow{2}{*}{\textsc{Relay} (sg)} & $\checkmark$ & 58 & 42\%{\scriptsize $\pm$\,0.11\%} & 7 & 62{\scriptsize $\pm$\,0.05} & 70 & 75\%{\scriptsize $\pm$\,0.92\%} & 6 & 13{\scriptsize $\pm$\,0.09} \\
    & $\times$ & 59 & 45\%{\scriptsize $\pm$\,3.06\%} & 7 & 59{\scriptsize $\pm$\,0.11} & 70 & 65\%{\scriptsize $\pm$\,3.18\%} & 6 & 01{\scriptsize $\pm$\,0.03} \\
    \midrule
    \multirow{2}{*}{\textsc{Relay}} & $\checkmark$ & \textbf{62} & \textbf{67\%}{\scriptsize $\pm$\,2.40\%} & \underline{7} & \underline{43}{\scriptsize $\pm$\,0.18} & \textbf{76} & \textbf{42\%}{\scriptsize $\pm$\,2.47\%} & \underline{5} & \underline{86}{\scriptsize $\pm$\,0.08} \\
    & $\times$ & \underline{62} & \underline{07\%}{\scriptsize $\pm$\,0.70\%} & \textbf{7} & \textbf{20}{\scriptsize $\pm$\,0.18} & \underline{73} & \underline{10\%}{\scriptsize $\pm$\,2.64\%} & \textbf{5} & \textbf{80}{\scriptsize $\pm$\,0.08} \\
    \bottomrule
  \end{tabular}
  }
  \caption{Sudoku exact match and mean NFE at $\tau = 0.15$.
    \emph{Unfiltered} reports performance on puzzles iterated from the test split in dataset order; \emph{deduction-only} restricts to puzzles whose solver trace requires Advanced/Master heuristics (no recursive backtracking).
    Accuracies are \% exact match, with sample s.d.\ across 3 training seeds. See \Cref{sec:dataset_details} for more details.}
  \label{tab:sudoku-tau015}
\end{table*}

\subsection{Pretrained Model Adaptation: Fast-dLLM v2}\label{sec:pretrained-model-adaptation}

Next, we investigate whether state-of-the-art DLMs can be efficiently adapted into \OursShort{} diffusion models with a limited amount of finetuning, and whether this adaptation can improve their accuracy-latency frontiers. 

\looseness=-1
\xhdr{Base model} As our base model, we choose Fast-dLLM v2 (1.5B parameters) \citep{wu2025fastdllmv2efficientblockdiffusion}, a state-of-the-art DLM adapted from Qwen2.5 \citep{qwen2025qwen25technicalreport} by finetuning on the LLaMA-Nemotron dataset \citep{bercovich2025llama}.

\xhdr{Training} For \OursShort{} adaptation we apply supervised fine-tuning to all parameters for $200$ optimizer steps at effective batch size $64$ on a 60{,}000-example mixture of filtered OpenCodeInstruct and OpenMathInstruct-2 examples with a $40{/}60$ code/math proportion (dataset and hardware details in \Cref{sec:fastdllm_training_setup}). To make \Cref{alg:training} compatible with state-of-the-art DLMs that combine block-autoregressive decoding with KV caching, we make two careful adaptations to the on-policy rollout. First, we run the $K{=}2$ relay rollout \emph{only inside the active block} of Fast-dLLM v2's BD3-LM-style doubled (block-causal $\oplus$ block-bidirectional) attention~\citep{arriola_block_2024,wu2025fastdllmv2efficientblockdiffusion}, leaving previously decoded blocks frozen so their inter-block KV cache is reused unchanged across both passes. Second, within the active block we update the relay state $\vh$ \emph{only at positions that are still masked}: clean (already-committed) sub-block tokens contribute attention but their relay entries are not overwritten, which keeps within-block sub-block KV cache entries valid as the block fills in. %Under matched batch and step budgets, \OursShort{} takes roughly $4.5\times$ the wall time of vanilla SFT, attributable to its second forward and BPTT through the relay.

\begin{table*}[t]
\centering
\footnotesize
\setlength{\tabcolsep}{16pt}
\renewcommand{\arraystretch}{1.12}
\begin{tabular}{lcccccc}
\toprule
\textbf{Method}
& \multicolumn{3}{c}{\textbf{HumanEval}}
& \multicolumn{3}{c}{\textbf{MBPP}} \\
\cmidrule(lr){2-4}\cmidrule(l){5-7}
& Base$\uparrow$ & Plus$\uparrow$ & NFE$\downarrow$
& Base$\uparrow$ & Plus$\uparrow$ & NFE$\downarrow$ \\
\midrule
\textbf{Fast-dLLM-v2 (1.5B)}
& $38.4\%$ & $35.4\%$ & $178.1$
& $46.8\%$ & $39.7\%$ & $133.0$ \\
\midrule
\quad Vanilla SFT 
& $38.4\%$ & $34.1\%$ & $130.7$
& $\underline{43.9\%}$ & $38.1\%$ & $84.8$ \\
\quad Rollout
& $\textbf{42.7\%}$ & $\textbf{39.6\%}$ & $118.4$
& $\textbf{46.6\%}$ & $\underline{39.9\%}$ & $\textbf{76.6}$ \\
\quad \OursShort{} (sg)
& $38.4\%$ & $35.4\%$ & $\underline{104.4}$
& $43.1\%$ & $39.2\%$ & $80.1$ \\
\quad \OursShort{}
& $\underline{42.1\%}$ & $\underline{37.2\%}$ & $\textbf{88.3}$
& $\textbf{46.6\%}$ & $\textbf{41.5\%}$ & $\underline{78.8}$ \\
\bottomrule
\end{tabular}
\setlength{\abovecaptionskip}{3pt}%
\caption{Pretrained adaptation on Fast-dLLM-v2 (1.5B), evaluated at threshold $0.85$. Average NFE is computed as the mean per-example count of active denoising forward calls during batched sample generation, excluding prompt prefill and final cache-update next-token forwards. Bold (best) and underlined (second-best) values are selected among adapted rows only, excluding the off-the-shelf baseline.}
\label{tab:pretrained-adaptation-fast-dllm-v2-nfe}
\end{table*}

\xhdr{Evaluation} Inference follows Fast-dLLM v2's confidence-based parallel decoding~\citep{wu2025fastdllmv2efficientblockdiffusion}: within each block, the backbone applies the token-shift head so masked positions are read from the preceding token's logit row, samples are drawn with top-$p$ filtering ($p{=}0.95$, temperature $0$), and a position unmasks when the probability of its sampled token exceeds a confidence threshold~$\tau$, while the argmax masked position in each active sub-block is always unmasked so every forward makes progress. We use block length $32$, sub-block length $8$, and $\tau{=}0.85$ for all HumanEval/MBPP numbers below (including NFE in \Cref{tab:pretrained-adaptation-fast-dllm-v2-nfe}). The \emph{Plus} columns report HumanEval+ and MBPP+ from EvalPlus~\citep{liu2023evalplus}---expanded unit-test suites released with the EvalPlus framework\footnote{\url{https://github.com/evalplus/evalplus}}---in the same \emph{Base}/\emph{Plus} layout used for code results in \citet{wu2025fastdllmv2efficientblockdiffusion}.

\looseness=-1
As in Sudoku, \OursShort{} is Pareto-optimal on the accuracy-NFE frontier. While the Rollout baseline (on-policy training rollouts with the relay channel disabled) posts the highest raw HumanEval accuracies, \OursShort{} reaches comparable accuracy at $25\%$ fewer NFEs ($88.3$ vs.\ $118.4$) and matches or exceeds Rollout on MBPP. Notably, \textbf{on HumanEval, \OursShort{} even surpasses the vanilla SFT accuracy at 32\% fewer NFEs} ($88.3$ vs.\ $130.7$), demonstrating that \OursShort{} improves both accuracy and the number of denoising steps required to reach it.

\subsubsection{Training memory overhead}\label{sec:fastdllm-memory-overhead}

\begin{figure}[t]
\centering
\includegraphics[width=0.85\linewidth]{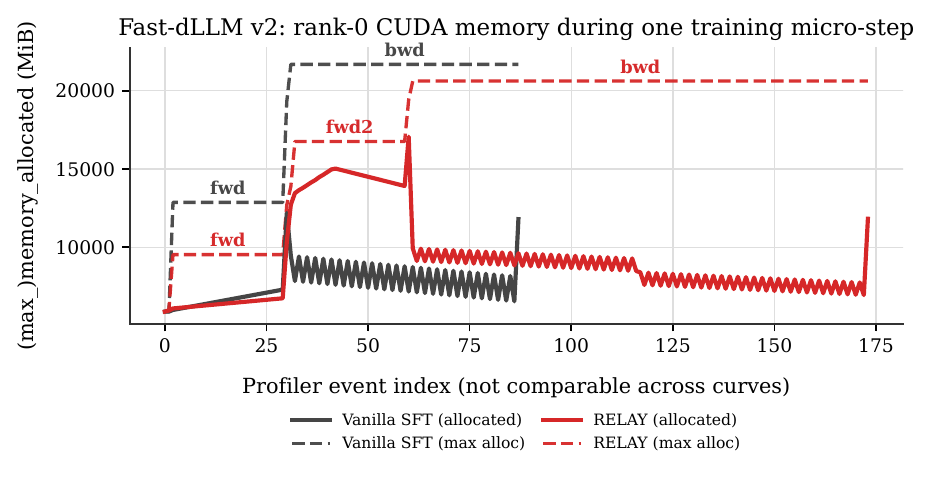}
\caption{GPU memory during one training micro-step of Fast-dLLM v2 on an A100~80GB. Solid lines show the live GPU memory at every decoder-layer forward/backward hook. Dashed lines show the running maximum of live memory within the same micro-step (high-water mark). Phase labels (\texttt{fwd}, \texttt{fwd2}, \texttt{bwd}) mark each phase's plateau. \OursShort{} carries higher live memory through \texttt{fwd2}, but its peak ($\approx\!20.1$\,GiB) lands within $\approx\!1$\,GiB of vanilla SFT's ($\approx\!21.2$\,GiB); see main text and \Cref{sec:fastdllm_memory_profiling} for the mechanism.}
\label{fig:fastdllm-memory-microstep}
\end{figure}

\looseness=-1
A natural concern is that BPTT through $K{=}2$ forward passes inflates training memory. \Cref{fig:fastdllm-memory-microstep} profiles one micro-step on an A100~80GB. Each regime is shown with two curves: the solid trace samples live GPU memory at every transformer-layer hook, and the dashed trace is its running maximum, a high-water mark whose final value is the peak the run actually demanded. Thus a short-lived allocation can lift the dashed trace even if it is freed before the next solid-line sample. The largest such transient---and the binding peak of the whole micro-step in both regimes---is the cross-entropy backward through the vocabulary-projection head (\texttt{lm\_head}), which materializes a $B\!\times\!T\!\times\!V$ fp32 grad-of-logits buffer at the start of \texttt{bwd}.

\OursShort{}'s second forward raises the live trace by $\approx\!5$\,GiB through \texttt{fwd2}: the saved activations of forward~1 and the relay state $\vh$ coexist with forward~2 to route credit through both passes (\Cref{alg:training}). Most of that elevation is autograd intermediates rather than saved-for-backward state, and PyTorch releases it in a single step before the \texttt{lm\_head} spike fires---live drops by $\approx\!7$\,GiB for \OursShort{} versus $\approx\!2.7$\,GiB for vanilla, leaving the two regimes within $\approx\!0.5$\,GiB of each other just before the spike. Adding the spike yields nearly identical peaks, $20.1$\,GiB for \OursShort{} versus $21.2$\,GiB for vanilla SFT---in fact, \OursShort{}'s larger pre-spike drop edges its peak slightly below vanilla's. This near-tie is structural rather than incidental: Fast-dLLM v2's vanilla SFT forward already doubles both the sequence and the batch (BD3-LM's $[\vx_t\,\|\,\vx_0]$ layout plus a complementary-mask copy along the batch), so each of \OursShort{}'s two forwards runs at half vanilla's per-pass batch and the two together demand memory comparable to vanilla's single doubled forward. BPTT through $K{=}2$ thus does not double peak memory in this setup (gradient checkpointing, ZeRO-3, non-fused CE)---\OursShort{} trades vanilla's in-forward batch doubling for an explicit second pass---and we expect peak memory to stay comparable whenever the vanilla baseline already pays a doubled-batch forward and the \texttt{lm\_head} backward dominates. Per-phase numbers and the profiling protocol are deferred to \Cref{sec:fastdllm_memory_profiling}.

\section{Related Work}\label{sec:related_work}

%\looseness=-1
%\xhdr{Masked diffusion language models}
Discrete diffusion models~\citep{austin_structured_2021}, which apply the iterative denoising principles of continuous diffusion~\citep{ho_ddpm_2020, song_score_2021} to categorical sequences, have emerged as a strong framework for language modeling. 
In particular, Masked Diffusion Models (MDMs)~\citep{sahoo_simple_2024,shi_simplified_2024}, which generate sequences by iterative unmasking have been shown to scale well to larger model sizes~\citep{nie_llada_2025, dream2025, wu2025fastdllm, wu2025fastdllmv2efficientblockdiffusion}.
%\looseness=-1
%\xhdr{Cross-step bottlenecks in denoising}
The same diffusion style training that makes MDMs simple also limits what can be communicated between denoising steps: rich internal representations are collapsed into sampled tokens before the next step begins. 
%\citet{hayakawa_distillation_2025} analyzes a related limitation of factored reverse transitions, showing that they may require many steps to approximate the true joint distribution and proposing mixture-model distillation as a remedy. 
Some recent works term the collapse of internal information as a ``sampling wall'' or ``information island''~\citep{jo_loopholing_2025, xia_metastate_2026}.

To address this, several recent approaches use a continuous relaxation or augmented state trajectories.
CADD~\citep{zheng_continuously_2025} pairs each discrete position with a continuous variable trained via a continuous diffusion process in the embedding space.
Soft-Masked Diffusion~\citep{hersche_softmasked_2025} passes output distributions or top-$k$ predictions from the previous step back into the input to the model for the next step.
%DRAKES~\citep{wang_fine-tuning_2025} proposes to backpropagates through discrete diffusion trajectories with a Gumbel-Softmax relaxation~\citep{jang_categorical_2017} in order to maximize a terminal reward. 
VADD~\citep{xie_variational_2025}, on the other hand, trains a VAE atop discrete diffusion.
All these approaches rely on leveraging a continuous diffusion process to carry more information across steps even though we ultimately only care about the discrete variables.
In contrast, our approach provides supervision through the discrete variables only.

MetaState~\citep{xia_metastate_2026}, Loopholing~\citep{jo_loopholing_2025}, and Residual Context Diffusion (RCD)~\citep{hu2026residual} introduce a continuous pathway that carries latent information across steps and similarly train it without relying on continuous diffusion.
MetaState adds a fixed-size working memory to frozen dLLMs and trains it over multi-step denoising rollouts. Loopholing and RCD, on the other hand, simply inject the hidden state or residual from the previous step into the input of the current step, like our \OursShort{} (sg) setting in the ablations. 
These methods allow the model to \emph{learn to use} the hidden state for future predictions. 
\OursShort{} goes one step further by training the hidden state end-to-end via BPTT which allows the model to also \emph{learn to generate} the hidden state for future predictions.

\section{Discussion}\label{sec:discussion}

\xhdr{Summary of results}
Masked diffusion models suffer from a \emph{hard reset} between denoising steps: the Transformer computes rich hidden states at every position---including those still masked---but discards them at the end of each forward pass, so the only information that persists is the discrete tokens just committed.
\OursShort{} addresses this by carrying the last-layer hidden states forward as a learned relay and training it end-to-end via truncated BPTT, so the model is explicitly rewarded for writing hidden states that will be useful to future denoising steps.
Empirically, the three components that constitute \OursShort{}---a rollout-based training procedure, passing the hidden state forward across denoising steps, and training the hidden state end-to-end via BPTT---each push the performance-latency frontier on their own, and combine constructively.
On Sudoku-Extreme, the full method attained the best accuracy-per-NFE point on the Pareto frontier (\Cref{fig:sudoku-pareto}); on Fast-dLLM v2 it outperformed standard supervised fine-tuning on coding tasks while reducing inference latency by up to $32\%$.

\xhdr{Limitations}
\OursShort{} introduces two computational trade-offs.
First, the relay mechanism adds a small per-step overhead for reading and writing the continuous relay state, though the reduced number of forward passes needed to reach a given accuracy can still yield a net inference-latency improvement.
Second, two-step BPTT during training increases live activation memory and per-step compute. In our Fast-dLLM v2 profile, however, each \OursShort{} forward runs at half the batch of vanilla SFT's doubled forward, so this added activation stays below the peak set by the vocabulary-projection (\texttt{lm\_head}) backward (\Cref{sec:fastdllm-memory-overhead}, \Cref{sec:fastdllm_memory_profiling}), leaving observed peak GPU memory nearly unchanged.
Overall, while \OursShort{} requires more training time than vanilla MLM training, this training-time gap can be amortized by improvements to the inference-time accuracy-latency frontier, and narrowed with more careful engineering.

\xhdr{Outlook and future work}
\OursShort{} is a meaningful step towards a non-greedy, forward-thinking approach to iterative non-autoregressive generation, and there are several natural follow-up directions.
The relay state gives a diffusion model a continuous substrate on which to carry intermediate computation across denoising steps; understanding what this state encodes, and whether it can be probed or steered, is a promising direction for interpreting and improving latent reasoning in MDMs.
Because the relay mechanism is largely architecture- and modality-agnostic, applying it beyond text---for example, to image or molecular discrete diffusion---is also a natural next step.

\section{Conclusion}\label{sec:conclusion}

We introduced \OursShort{} to address the hard reset problem in MDMs by passing a continuous, differentiable latent state across inference steps.
By training a relay channel via truncated BPTT, we demonstrated that discrete diffusion models can explicitly optimize intermediate representations for future unmasking decisions, advancing the performance-latency Pareto frontier.

\section*{Acknowledgments}
DP, BR, and AM thank Michael Boratko for helpful initial discussions.
DP and BR acknowledge support from IBM under IBM Research Collaboration Agreement No.\ W1668553 and from the National Science Foundation under grant IIS-2106391.
NB acknowledges support from an NSF Graduate Research Fellowship, Quad Fellowship, and Mercor Graduate Fellowship.
TGJR acknowledges support provided, in part, by the Province of Ontario, the Government of Canada through CIFAR, the Vector Institute for Artificial Intelligence, and by the Digital Research Alliance of Canada (\href{alliancecan.ca}{alliancecan.ca}).

% \begin{ack}
% Funding and competing interests disclosures.
% \end{ack}

\bibliographystyle{unsrtnat}
% Looser than the aggressive 0.12ex pass: spreads refs across three pages.
\setlength{\bibsep}{0.65ex plus 0.18ex minus 0.12ex}
\bibliography{references}

\clearpage

\begin{appendices}

\crefalias{section}{appsec}
\crefalias{subsection}{appsec}
\crefalias{subsubsection}{appsec}

\setcounter{equation}{0}
\renewcommand{\theequation}{\thesection.\arabic{equation}}

\setcounter{footnote}{0}
\setcounter{tocdepth}{3}

\section*{\LARGE Appendix}
\vskip 0.5em minus 0.2em
\phantomsection
\addcontentsline{toc}{section}{Appendix}

\tableofcontents
\vskip 1em plus 0.5em

\newpage
\section{Experimental Details}\label{sec:appendix_details}

\subsection{Sudoku}

\subsubsection{Dataset}\label{sec:dataset_details}

\xhdr{Sudoku Extreme}
We train and evaluate on our derived dataset built on top of Sudoku-Extreme~\citep{wang2025hierarchical} by running the solver of \citet{vink2024sudokusolver} over every puzzle.\footnote{Derived dataset: \url{https://huggingface.co/datasets/brozonoyer/sapientinc-sudoku-extreme-timvink-sudoku-solver}. Solver code: \url{https://github.com/timvink/sudoku-solver}.} The base dataset consists of 9$\times$9 Sudoku puzzles with 17 given clues, the minimum number compatible with a uniquely solvable puzzle~\citep{mcguire2014there}. Each puzzle is represented as a flat sequence of length $L{=}81$ over an alphabet of 11 task tokens: digits $\{1, \ldots, 9\}$, a blank/zero token for unfilled cells, and a mask token. The clue positions are treated as fixed and are not modified during inference; the remaining 64 positions are mutable. Our derived version augments each puzzle with:
\begin{itemize}[nosep,leftmargin=*]
  \item \texttt{trajectory}: step-by-step board states from question to solution
  \item \texttt{num\_steps}: number of solver calls to reach the solution
  \item \texttt{strategies\_used}: set of human-like deduction strategies invoked (used by the deduction-only cohort below)
\end{itemize}
We use the training split (3,831,994 puzzles) and evaluate on the test split (422,786 puzzles), and validate on the first 100 batches at batch size 512 (51,200 puzzles) per checkpoint.

\xhdr{Deduction-only cohort}\label{par:deduction-cohort}
For the qualitative legality analysis of \Cref{sec:sudoku} we use the \texttt{strategies\_used} field described above to filter puzzles. Since the studied architectures cannot perform recursive search, we keep only puzzles whose solver trace contains \emph{Advanced} (Naked Pair, Hidden Pair, Naked Triple, Hidden Triple, Naked Quad, Hidden Quad) or \emph{Master} (X-Wing, Swordfish, Jellyfish, Forcing Chain) strategies and never falls back on recursive backtracking. The resulting cohort contains 2{,}000 test puzzles (1{,}933 Advanced + 67 Master).

\xhdr{Evaluation protocol for \Cref{tab:sudoku-tau015}}
Each cell of \Cref{tab:sudoku-tau015} aggregates the first $N{=}2000$ puzzles from the Hugging Face test split in dataset order; for the deduction-only cohort, we keep the first 2{,}000 examples whose solver trace uses Advanced or Master strategies without recursive backtracking.

\subsubsection{Model Architecture}\label{subsec:model_architectures}

The backbone for all Sudoku experiments (\Cref{tab:sudoku-tau015}) is a shallow rotary Transformer:
\begin{itemize}[itemsep=0pt]
  \item \textbf{Depth / width:} $L{=}4$ layers, hidden dimension $d_{\mathrm{model}}{=}384$, feedforward width $4d_{\mathrm{model}}{=}1536$
  \item \textbf{Attention:} $H{=}6$ heads (head dimension $d_{\mathrm{model}}/H{=}64$), rotary positional embeddings (rotary width~$64$)
  \item \textbf{MLP:} ReLU nonlinearities, dropout $0.1$
  \item \textbf{Vocabulary:} the 11 task tokens described in \Cref{sec:dataset_details}, plus special tokens (6 reserved), giving the embedding table vocabulary size $|\mathcal{V}|{=}17$
\end{itemize}

The \textsc{Relay} variant adds a differentiable carry channel following~\citet{jo_loopholing_2025}. At each inference step the relay tensor $h_t$ from the previous step is normalized by an affine LayerNorm ($\varepsilon_{\mathrm{LN}}{=}10^{-5}$), yielding $\delta_t = \mathrm{LN}_{\mathrm{relay}}(h_t)$, and injected additively into the residual stream before layer zero: $x \leftarrow \mathrm{Embed}(x_t)+\delta_t$. The outgoing relay state $h_{t+1}$ is read from the final transformer block, while logits are always produced from the same terminal hidden states. We initialize $\mathrm{LN}_{\mathrm{relay}}$ with PyTorch defaults ($\boldsymbol{\gamma}_{\mathrm{relay}} \gets \mathbf{1}$, $\boldsymbol{\beta}_{\mathrm{relay}} \gets \mathbf{0}$).
We implement all the models using the xLM~\citep{patel-etal-2026-xlm} package, which provides a unified interface for training and inference of non-autoregressive language models making the ablations and experiments easy to reproduce.

Parameter counts (with and without weight tying) are:
\begin{itemize}[itemsep=0pt]
  \item \textbf{Baseline} (MLM / rollout-buffer only): 7{,}105{,}536 untied; 7{,}099{,}008 tied
  \item \textbf{\textsc{Relay}}: 7{,}106{,}304 untied; 7{,}099{,}776 tied
\end{itemize}

\subsubsection{Training Hyperparameters}

\begin{itemize}[itemsep=0pt]
  \item \textbf{Batch size:} 512 (single GPU, bf16 mixed precision)
  \item \textbf{Optimizer:} AdamW, learning rate $5\times10^{-4}$, weight decay $10^{-2}$
  \item \textbf{LR schedule:} constant with 2{,}000-step linear warmup, no decay thereafter
  \item \textbf{Gradient clipping:} global Frobenius norm $0.5$
  \item \textbf{BPTT unroll horizon:} $K{=}2$ steps (\textsc{Relay} runs only)
  \item \textbf{Confidence threshold:} $\tau{=}0.15$ (maximum softmax probability), perturbed by $\mathcal{N}(0,0.1^2)$ during training and fixed at inference
  \item \textbf{Validation:} every 5{,}000 steps on 100 batches; threshold sweep $\tau\in\{0.05,0.10,\ldots,0.25\}$
  \item \textbf{Total steps:} 300{,}000; results reported in \Cref{tab:sudoku-tau015}
\end{itemize}

\subsection{Fast-dLLM v2}

\subsubsection{Dataset}

\paragraph{OpenCode/OpenMath c40m60 mixture for Fast-dLLM v2 adaption.}
For Fast-dLLM v2 adaptation, we use a 60k-example supervised fine-tuning mixture from \texttt{nvidia/OpenCodeInstruct}\footnote{\url{https://huggingface.co/datasets/nvidia/OpenCodeInstruct}} and \texttt{nvidia/OpenMathInstruct-2}\footnote{\url{https://huggingface.co/datasets/nvidia/OpenMathInstruct-2}}, with 24k code examples and 36k math examples. We filter for high-quality prompt--answer pairs, remove held-out evaluation contamination, format examples as one-turn conversations, and cap sequences at 2048 tokens.

\subsubsection{Model Architecture}

Unlike standard MDMs, which denoise the entire token sequence globally, Fast-dLLM v2 models \citep{wu2025fastdllmv2efficientblockdiffusion} a block-wise Markov process. By partitioning the sequence into blocks of size $D$, it targets the local conditional distribution $p_\theta(x^b | x_t^b, x_0^{<b})$. This localizes the diffusion process while anchoring it to an autoregressive prefix, successfully bypassing the immense pretraining costs associated with full-attention MDMs.

The core architectural shift lies in its attention topology. Fast-dLLM v2 concatenates the noised $x_t$ and clean $x_0$ sequences into a $2L$-length tensor, governed by a full attention mask $\mathcal{M}_{full} \in \{0, 1\}^{2L \times 2L}$ \citep{arriola_block_2024}. This mask explicitly splits into three distinct functional roles:
\begin{itemize}[itemsep=0pt]
    \item $\mathcal{M}_{BD}$: Enables intra-block bidirectional attention within each block.
    \item $\mathcal{M}_{OBC}$: Allows the noised block to attend to the completely denoised, clean prefix $x_0^{<b}$.
    \item $\mathcal{M}_{BC}$: Enforces standard left-to-right causality among the clean tokens.
\end{itemize}
The $2L$ concatenation lets the noised and clean views be processed in a single forward pass. On top of this, a complementary masking strategy trains on both a sampled mask $m$ and its complement $\bar{m} = 1 - m$, so that every token in the input contributes supervision rather than only those masked under $m$.

At inference, this topology enables hierarchical Key-Value caching—a major advantage over standard MDMs, which typically require full-sequence recomputation at every denoising step. Completely denoised blocks $x_0^{<b}$ are saved as read-only context, while a DualCache handles prefix and suffix activations within the active, partially noised block $x_t^b$.

\subsubsection{Fast-dLLM v2 training hardware and parallelism}\label{sec:fastdllm_training_setup}
All adaptation runs use DeepSpeed ZeRO-3 with bf16 mixed precision and gradient checkpointing on two NVIDIA A100~80GB GPUs, with per-device batch size $2$ and gradient accumulation $16$ (effective batch size $64$). For \OursShort{} adaptation, unlike for the Sudoku experiments, $\mathrm{LN}_{\mathrm{relay}}$ uses zero-initialized $\boldsymbol{\gamma}_{\mathrm{relay}}$ (with $\boldsymbol{\beta}_{\mathrm{relay}}{=}0$), so early forward passes approximate an identity relay until training updates $\boldsymbol{\gamma}_{\mathrm{relay}}$.

\section{Fast-dLLM v2 memory profiling}\label{sec:fastdllm_memory_profiling}

This section gives the protocol and per-phase numbers behind \Cref{fig:fastdllm-memory-microstep}, repeated below.

\begin{center}
\includegraphics[width=0.80\linewidth]{figures/gpu_memory_vanilla_vs_relay_c40m60_rank0.pdf}
\end{center}

\xhdr{Setup} We profile a single training micro-step of Fast-dLLM v2 on the OpenCode/OpenMath \texttt{c40m60} mixture under the same hardware and parallelism as the main runs (\Cref{sec:fastdllm_training_setup}): two A100~80GB GPUs, DeepSpeed ZeRO-3, bf16, and gradient checkpointing, with sequence length $2048$ and per-device batch size $2$. Production runs use gradient accumulation $16$; profiling forces accumulation to $1$ and replaces \texttt{optimizer.step} with a no-op so that the recorded peak is attributable to a single forward/backward pair rather than to optimizer-state allocation.

\xhdr{Instrumentation} On every decoder-layer forward and backward hook we log \texttt{memory\_allocated} and \texttt{max\_memory\_allocated} from \texttt{torch.cuda}---the solid and dashed traces in \Cref{fig:fastdllm-memory-microstep}, respectively; we call \texttt{reset\_peak\_memory\_stats()} once at the start of the profiled micro-step so the dashed series is a within-step high-water mark rather than a long-run accumulator. All measurements are taken in eager mode with \texttt{torch.compile} and FlashAttention~2 disabled, so steps in the dashed roof correspond directly to discrete kernel-level allocations. Phase labels (\texttt{fwd}, \texttt{fwd2}, \texttt{bwd}) are placed at each phase's plateau in the dashed series. Horizontal axes are profiler event indices ($88$ for vanilla SFT, $174$ for \OursShort{}) and are not directly comparable across the two curves.

\xhdr{Per-phase peaks} The dashed all-time peaks settle at $20{,}618$\,MiB ($\approx\!20.1$\,GiB) for \OursShort{} versus $21{,}683$\,MiB ($\approx\!21.2$\,GiB) for vanilla SFT. In both regimes the peak is set at the start of \texttt{bwd}, when the cross-entropy backward through \texttt{lm\_head} transiently allocates a $B\!\times\!T\!\times\!V$ gradient-of-logits buffer that HuggingFace materializes in fp32 for numerical stability, on top of the bf16 logits tensor it is differentiating. The effective batch $B$ is not the same in the two regimes: Fast-dLLM v2's vanilla SFT forward applies BD3-LM's complementary-mask doubling along the batch dimension---in addition to the $[\vx_t\,\|\,\vx_0]$ doubling along the sequence dimension that both regimes share---whereas each of \OursShort{}'s two rollout forwards runs at the undoubled batch. At per-device batch size $2$ (with $T{=}2048$, $V{=}151936$), the fp32 buffer is therefore $[4, 2048, 151936]\!\times\!4\,\text{B}\!\approx\!4.6$\,GiB for vanilla but only half that, $\approx\!2.3$\,GiB, for \OursShort{}. \OursShort{} offsets this smaller spike by retaining \emph{two} forwards' activations: its second forward elevates the live trace by $\approx\!5$\,GiB through \texttt{fwd2}---saved activations of forward~1 plus the relay state $\vh$ must coexist with forward~2 to provide credit through both passes (\Cref{alg:training})---reaching $17{,}057$\,MiB at \texttt{relay\_fwd2\_end}, still below the $20{,}618$\,MiB peak that the CE backward sets one event later, so the live plateau through \texttt{fwd2} does not become the binding peak. These two asymmetries---vanilla's larger CE buffer versus \OursShort{}'s heavier retained activations---roughly cancel, leaving the peaks within $\approx\!1$\,GiB. The residual gap in \OursShort{}'s favor is structural rather than allocator noise: more autograd intermediates from forward~1 are released by the time the CE backward fires than vanilla releases by the analogous event ($\Delta\text{live}\!=\!-7{,}119$\,MiB across this transition for \OursShort{} vs.\ $-2{,}727$\,MiB for vanilla). Rank-$1$ traces reproduce both peaks within $<\!0.1$\,MiB, so \Cref{fig:fastdllm-memory-microstep} shows only rank~$0$.

\section{Deeper BPTT Ablations}\label{sec:bptt_horizon}

We ablate the truncated-BPTT unroll horizon $K \in \{2, 3, 4\}$ for \OursShort{} on Sudoku-Extreme, keeping all other training hyperparameters as in \Cref{sec:appendix_details}.

\xhdr{Longer horizons do not improve performance}
Final exact match on a large 51,200-puzzle validation prefix is flat through $K=4$ (\Cref{tab:bptt_horizon_perf}). On the 2,000-puzzle decoding protocol of \Cref{tab:sudoku-tau015}, mean NFE stays at about 7-8 steps for every $K$. A short backpropagation window is sufficient for solve quality, and unrolling further increases training memory requirements without improving temporal credit assignment.

\xhdr{Optimization becomes less reliable at longer horizons}
\Cref{tab:bptt_horizon_stability} reports the pre-clip global gradient norm, together with the norm of the relay state and of its gradient, on runs that finish training. The relay state neither explodes nor vanishes. Pre-clip gradient norms, however, grow roughly threefold from K=2 to K=4 with the largest variance at K=4, and one of six K=4 runs (tied embeddings, seed 3) diverged and is excluded from Tables \ref{tab:bptt_horizon_perf} and \ref{tab:bptt_horizon_stability}.

\begin{table}[h]
  \centering
  \footnotesize
  \begin{tabular}{cccc}
    \toprule
    $K$ & Exact match (\%) & NFE \\
    \midrule
    2 & $64.8 \pm 1.0$ & $7.12 \pm 0.15$ \\
    3 & $64.4 \pm 0.6$ & $7.46 \pm 0.29$ \\
    4 & $63.6 \pm 1.3$ & $7.21 \pm 0.12$ \\
    \bottomrule
  \end{tabular}
  \caption{\textbf{BPTT horizon vs.\ Sudoku-Extreme performance.} Exact match is the final in-training validation on a large 51,200-puzzle prefix, pooled over tied and untied embeddings (mean $\pm$ s.d., 6 runs for $K=2,3$ and 5 runs for $K=4$). NFE is the 2,000-puzzle protocol of \Cref{tab:sudoku-tau015}, pooled over a single seed of the tied and untied settings.}
  \label{tab:bptt_horizon_perf}
\end{table}

\begin{table}[h]
  \centering
  \footnotesize
  \begin{tabular}{lcccccc}
    \toprule
    & $K$ & @10k & @50k & @100k & @200k & @300k \\
    \midrule
    \multirow{3}{*}{Pre-clip global gradient norm}
    & 2 & $0.54 \pm 0.13$ & $0.39 \pm 0.04$ & $0.39 \pm 0.04$ & $0.54 \pm 0.04$ & $0.56 \pm 0.06$ \\
    & 3 & $1.00 \pm 0.16$ & $0.75 \pm 0.04$ & $0.77 \pm 0.08$ & $0.99 \pm 0.12$ & $0.93 \pm 0.09$ \\
    & 4 & $1.49 \pm 0.19$ & $1.39 \pm 0.21$ & $1.32 \pm 0.13$ & $1.90 \pm 0.51$ & $1.66 \pm 0.19$ \\
    \midrule
    \multirow{3}{*}{Relay-state norm $\|\vh\|_2$}
    & 2 & $6.3 \pm 1.0$ & $10.5 \pm 1.3$ & $12.8 \pm 1.9$ & $14.4 \pm 1.9$ & $14.9 \pm 1.9$ \\
    & 3 & $6.0 \pm 0.9$ & $10.4 \pm 0.8$ & $12.3 \pm 1.3$ & $13.4 \pm 1.3$ & $13.6 \pm 1.3$ \\
    & 4 & $5.8 \pm 0.2$ & $10.4 \pm 0.5$ & $12.1 \pm 0.3$ & $12.7 \pm 0.4$ & $12.3 \pm 0.5$ \\
    \midrule
    \multirow{3}{*}{Per-step relay gradient norm $\|\nabla_{\vh}\|_2$}
    & 2 & $3.8 \pm 0.7$ & $7.5 \pm 1.2$ & $9.4 \pm 1.7$ & $11.1 \pm 1.9$ & $11.8 \pm 1.6$ \\
    & 3 & $3.7 \pm 0.6$ & $8.2 \pm 0.5$ & $9.9 \pm 1.2$ & $11.0 \pm 1.2$ & $11.6 \pm 1.1$ \\
    & 4 & $4.2 \pm 0.2$ & $8.4 \pm 0.7$ & $9.3 \pm 0.3$ & $10.9 \pm 0.5$ & $10.6 \pm 0.8$ \\
    \bottomrule
  \end{tabular}
  \caption{\textbf{Training-stability diagnostics across BPTT horizons $K$, measured at 10k--300k optimizer steps.} Mean $\pm$ s.d. over the same completed runs as \Cref{tab:bptt_horizon_perf}'s exact-match column.}
  \label{tab:bptt_horizon_stability}
\end{table}

\end{appendices}

\end{document}